%% file: main.tex
\documentclass[runningheads]{llncs}
\usepackage[T1]{fontenc}
\usepackage{graphicx,verbatim}
\usepackage{hyperref}
\usepackage{color}

\usepackage{caption}
\usepackage{subcaption}
\usepackage{amsmath}
\usepackage{amssymb}
\usepackage{tabularx}

\definecolor{gray}{rgb}{0.4,0.4,0.4}

\begin{document}
\title{Boosting 3D Liver Shape Datasets with Diffusion Models and Implicit Neural Representations}
\titlerunning{3D Liver HyperDiffusion}
%



\author{Khoa Tuan Nguyen\inst{1,2} \and
Francesca Tozzi\inst{4,6} \and
Wouter Willaert\inst{4,6} \and \\
Joris Vankerschaver\inst{2,3} \and
Nikdokht Rashidian\inst{5,6} \and
Wesley De Neve\inst{1,2}}

\authorrunning{Khoa et al.}

\institute{
IDLab, ELIS, Ghent University, Ghent, Belgium \and
Center for Biosystems and Biotech Data Science, Ghent University Global Campus, Incheon, Korea \and
Department of Applied Mathematics, Informatics, and Statistics, Ghent University, Ghent, Belgium \\
\email{\{khoatuan.nguyen,joris.vankerschaver,wesley.deneve\}@ghent.ac.kr} \and 
Department of GI Surgery, Ghent University Hospital, Ghent, Belgium \and
Department of HPB Surgery \& Liver Transplantation, Ghent University Hospital, Ghent, Belgium \and
Department of Human Structure and Repair, Ghent University, Ghent, Belgium \\
\email{\{francesca.tozzi,nikdokht.rashidian,wouter.willaert\}@ugent.be}
}

\maketitle              
\begin{abstract}
While the availability of open 3D medical shape datasets is increasing, offering substantial benefits to the research community, we have found that many of these datasets are, unfortunately, disorganized and contain artifacts.
These issues limit the development and training of robust models, particularly for accurate 3D reconstruction tasks.  
In this paper, we examine the current state of available 3D liver shape datasets and propose a solution using diffusion models combined with implicit neural representations (INRs) to augment and expand existing datasets.  
Our approach utilizes the generative capabilities of diffusion models to create realistic, diverse 3D liver shapes, capturing a wide range of anatomical variations and addressing the problem of data scarcity.  
Experimental results indicate that our method enhances dataset diversity, providing a scalable solution to improve the accuracy and reliability of 3D liver reconstruction and generation in medical applications. 
Finally, we suggest that diffusion models can also be applied to other downstream tasks in 3D medical imaging.



\end{abstract}
%
%
%

\input{chapters/1_Introduction}

\input{chapters/2_Dataset}

\input{chapters/3_Method}

\input{chapters/4_Experiment}

\input{chapters/5_Conclusions}

%
%
%
\bibliographystyle{splncs04}
\bibliography{ref}
\end{document}

%% file: chapters/1_Introduction.tex
\section{Introduction}

\textbf{Motivation.}
We are focused on 3D generation of human abdominal organs, particularly the liver, and during the preparation of the 3D liver dataset, we discovered that pretrained diffusion models~\cite{lee2024text} cannot be utilized because their training datasets~\cite{deitke2024objaverse,schuhmann2022laion} lack organ-specific information. 
Thus, we need to train from scratch using 3D medical shape datasets~\cite{li2023medshapenet,li2024abdomenatlas,montana2024saramis}.  
However, after analysis, we found that these datasets cannot be fully utilized due to the presence of artifacts.  
To address the problem of insufficient 3D objects\footnote{We use \textbf{`object'} to distinguish the 3D model from the term \textbf{`model'} in deep learning.}, we propose a new unconditional 3D liver HyperDiffusion model to generate synthetic datasets that supplement the limited amount of available real liver data.

\textbf{Implicit Neural Representations (INRs).}
INRs have recently emerged as a robust method for encoding various continuous functions, including images, video sequences, audio sequences, and 3D objects, utilizing a Multi-Layer Perceptron (MLP) for representation learning~\cite{Occupancy_Networks,sitzmann2019siren,tancik2020fourfeat}.  
When applied to complex 3D objects, INRs overcome the limitations of conventional discrete representations, such as point clouds or meshes~\cite{deluigi2023inr2vec}.
However, INRs lack generalizability because MLPs tend to suffer from overfitting to individual objects, making it impractical to train a new MLP for each new object.

\textbf{HyperNetworks.}
The use of a HyperNetwork~\cite{ha2017hypernetworks} refers to the use of an approach that leverages one network (called a HyperNetwork) to generate the weights for another network (called the main network), which eliminates the need to train the main network from scratch. 
Recent studies have demonstrated the application of this method in optimizing weights~\cite{Peebles2022}, generating audio samples~\cite{szatkowski2023hypernetworks}, and constructing 3D~\cite{erkocc2023hyperdiffusion,jun2023shap} generative models.  
Recognizing the potential of this approach, we utilize HyperDiffusion~\cite{erkocc2023hyperdiffusion}, a denoising diffusion model~\cite{ho2020denoising} designed to generate the weights of a MLP that represents a 3D object using the INR paradigm.  
We apply this model to address the generalization problem in INRs and obtain accurate synthetic 3D liver objects from the generated MLP.

\textbf{Our Contributions.}
In this paper, we analyze various overlooked issues in 3D medical shape datasets and propose an approach to supplement the limited amount of available 3D liver data by creating a hybrid 3D liver dataset that combines high-quality real 3D liver objects with synthetic ones.
As part of our validation, we conducted a survey where experts classified most of our synthesized 3D liver objects as real, demonstrating the quality and fidelity of our synthetic dataset.

Currently, few, if any, detailed investigations into the underlying causes of 3D liver generative model phenomena have been published.  
The most closely related study sharing a similar use case with ours is the recent work on TrIND~\cite{sinha2024trind}, which focuses on anatomical trees, whereas our work concentrates on abdominal organs, particularly the liver.

%% file: chapters/2_Dataset.tex
\begin{figure}[t]
    \centering
    \begin{subfigure}[b]{0.27\textwidth}
        \centering
        \includegraphics[width=\textwidth]{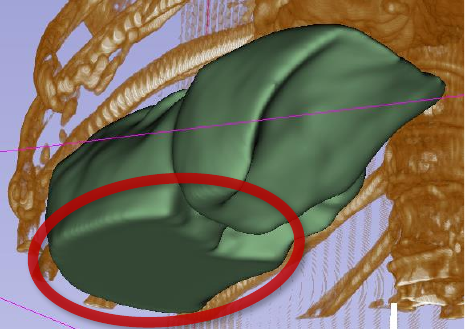}
        \caption{No full shape}
        \label{fig:TotalSegmentator_a}
    \end{subfigure}
    \hfill
    \begin{subfigure}[b]{0.65\textwidth}
        \centering
        \includegraphics[width=\textwidth]{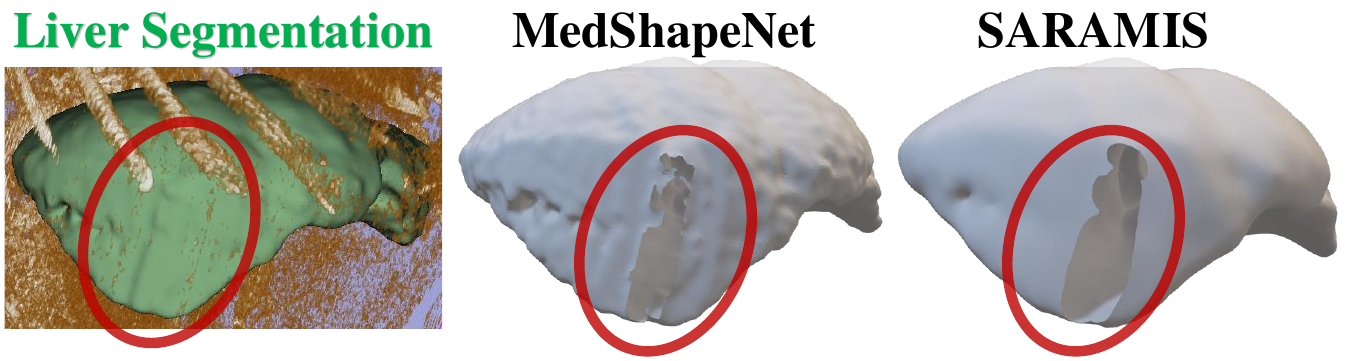}
        \caption{Artifacts from segmentation-to-3D conversion}
        \label{fig:TotalSegmentator_b}
    \end{subfigure}
    \vfill
    \begin{subfigure}[b]{1\textwidth}
        \centering
        \includegraphics[width=\textwidth]{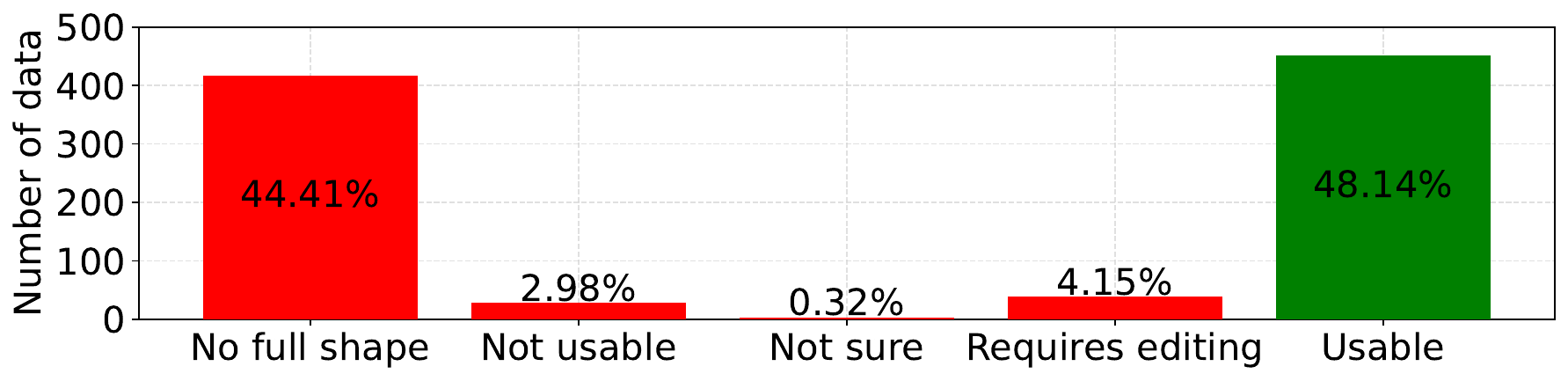}
        \caption{Percentage of TotalSegmentator (939 livers) by status}
        \label{fig:TotalSegmentator_c}
    \end{subfigure}
    \caption{
        Our analysis of TotalSegmentator, MedShapeNet, and SARAMIS shows that current 3D medical shape datasets are largely built on 3D segmentation results, such as those from TotalSegmentator.  
        This approach results in disorganized datasets, with usable liver objects (labeled `Usable' as a \textcolor{green}{green bar}) accounting for only $48.14\%$, while the others (labeled `No full shape', `Not usable', `Not sure', and `Requires editing' as \textcolor{red}{red bars}) are prone to artifacts and largely unusable in the case of TotalSegmentator.
    }
    \label{fig:TotalSegmentator}
\end{figure}
 
\section{Dataset Analysis}
\label{sec:data_analysis}

Currently, the Laion-5b~\cite{schuhmann2022laion} and Objaverse-XL~\cite{deitke2024objaverse} datasets are widely used to train text-image or text-3D pair diffusion models at scale~\cite{po2024state}.  
Objaverse-XL, which contains over 10 million 3D objects, is the largest dataset in terms of both scale and diversity, driving recent advancements in 3D generation~\cite{lee2024text,po2024state} and 3D reconstruction~\cite{yunus2024recent}.  
However, it lacks real 3D medical shapes for applications in the medical field. 
Fortunately, several open 3D medical shape datasets have been released, including SARAMIS~\cite{montana2024saramis}, MedShapeNet~\cite{li2023medshapenet}, and AbdomenAtlas~\cite{li2024abdomenatlas}.
Upon analysis, we found that these datasets generate 3D objects based on 3D organ segmentations from CT scans, with the most common source being the TotalSegmentator dataset~\cite{wasserthal2023totalsegmentator}.  
Therefore, in this study, we analyzed the TotalSegmentator dataset in more detail and identified the following problems, as shown in Fig.~\ref{fig:TotalSegmentator}.

The first problem is that the 3D objects are incomplete.
Since 3D organ segmentation does not require complete shapes, the resulting objects are often missing parts, which we refer to as `No full shape'.
For instance, the liver $s0046$ in Fig.~\ref{fig:TotalSegmentator_a} is incomplete, missing the lower part due to the CT scan being taken in the thoracic region.

The second problem is the presence of artifacts during the segmentation-to-3D conversion process.
Even when the 3D segmentation is correct, the exported 3D object can have artifacts, such as holes, as shown by liver $s0095$ in Fig.~\ref{fig:TotalSegmentator_b}.
After resolving the artifacts through a proper segmentation-to-3D conversion in 3D Slicer\footnote{\url{https://www.slicer.org}}\cite{fedorov20123d}, we requested our collaborating surgical team at **** Hospital to manually review 939 liver objects exported from 1228 CT scan subjects in the TotalSegmentator dataset and categorize them by status.
They determined that only $48.14\%$ (452 objects) are usable (`Usable'), while the remaining are categorized as follows: $44.41\%$ (417 objects) as `No full shape', $2.98\%$ (28 objects) as `Not usable' due to not resembling a real liver, $0.32\%$ (3 objects) as `Not sure', and $4.15\%$ (39 objects) as `Requires editing', such as cases where segmentation errors caused two organs to merge, needing re-segmentation or manual separation.
Fig.~\ref{fig:TotalSegmentator_c} shows the overall distribution.

With the goal of providing high-quality 3D objects from these 3D medical shape datasets, the problems we identified (which may not be limited to the TotalSegmentator dataset but could also apply to other CT datasets) can lead to errors during use.
For example, in the 3D liver generation task, generated livers may not be in full shape due to the dataset containing `No full shape' objects.
Therefore, we follow recent works~\cite{bluethgen2024vision,ghalebikesabi2023differentially,zhang2024diffboost} to address current dataset insufficiencies and propose generating synthetic datasets, either as 3D object datasets or INR datasets~\cite{maimplicit}, rather than investing time in analyzing additional datasets.


%% file: chapters/3_Method.tex
\section{Synthetic Dataset Generation}

\begin{figure}[t]
    \centering
    \begin{subfigure}[b]{0.34\textwidth}
        \centering
        \includegraphics[width=\textwidth]{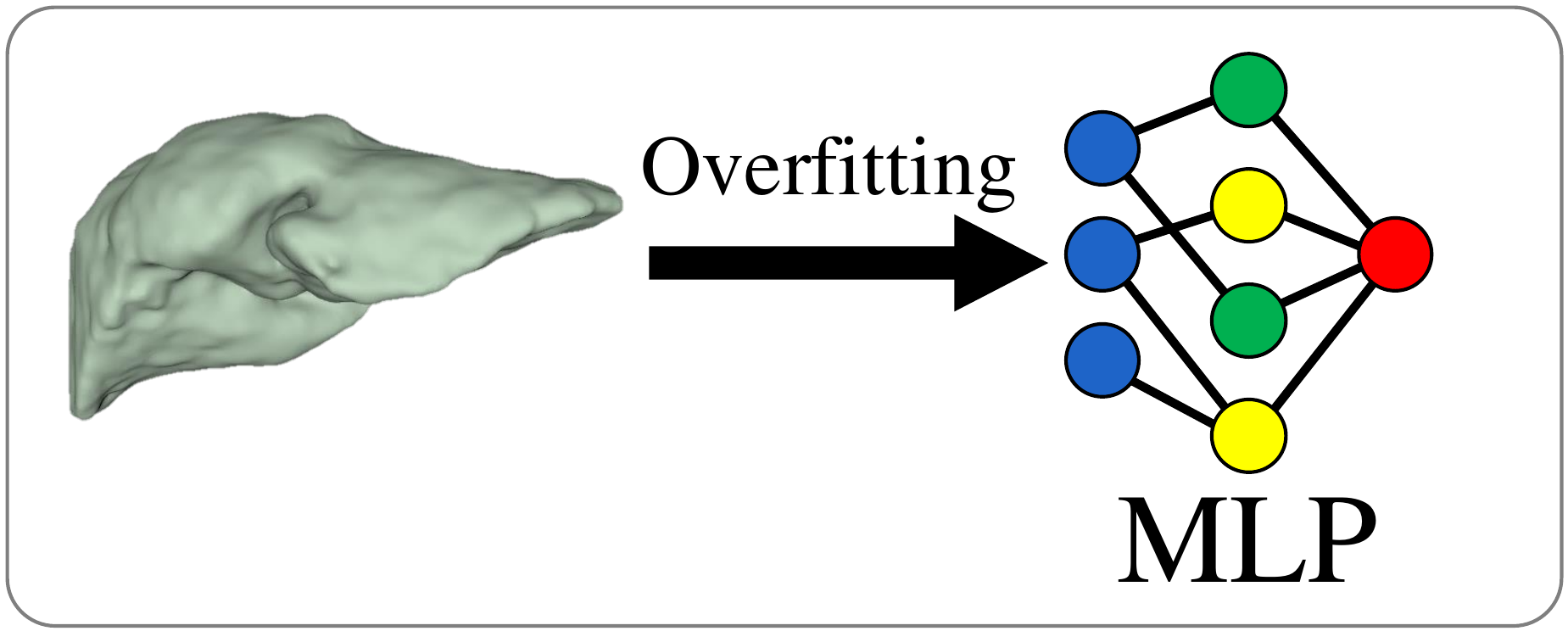}
        \caption{MLP training}
        \label{fig:overview_1}
    \end{subfigure}
    \hfill
    \begin{subfigure}[b]{0.58\textwidth}
        \centering
        \includegraphics[width=\textwidth]{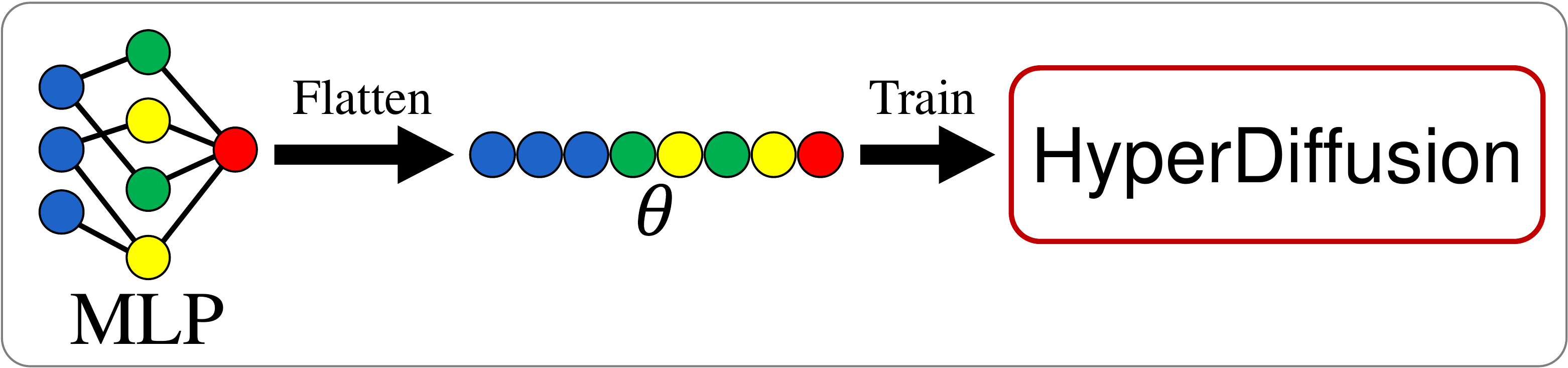}
        \caption{HyperDiffusion on MLP training}
        \label{fig:overview_2}
    \end{subfigure}
    \vfill
    \begin{subfigure}[b]{1\textwidth}
        \centering
        \includegraphics[width=\textwidth]{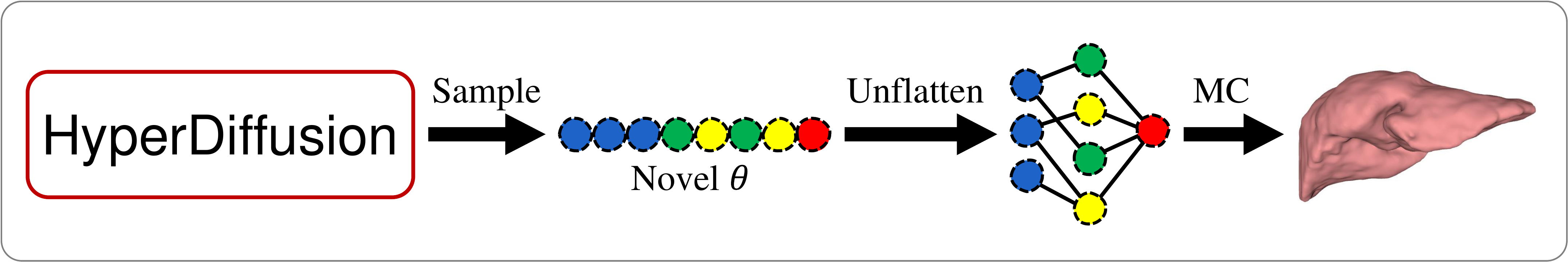}
        \caption{Synthesizing a novel MLP and reconstructing a new 3D liver object from it}
        \label{fig:overview_3}
    \end{subfigure}
    \vfill
    \begin{subfigure}[b]{1\textwidth}
        \centering
        \includegraphics[width=\textwidth]{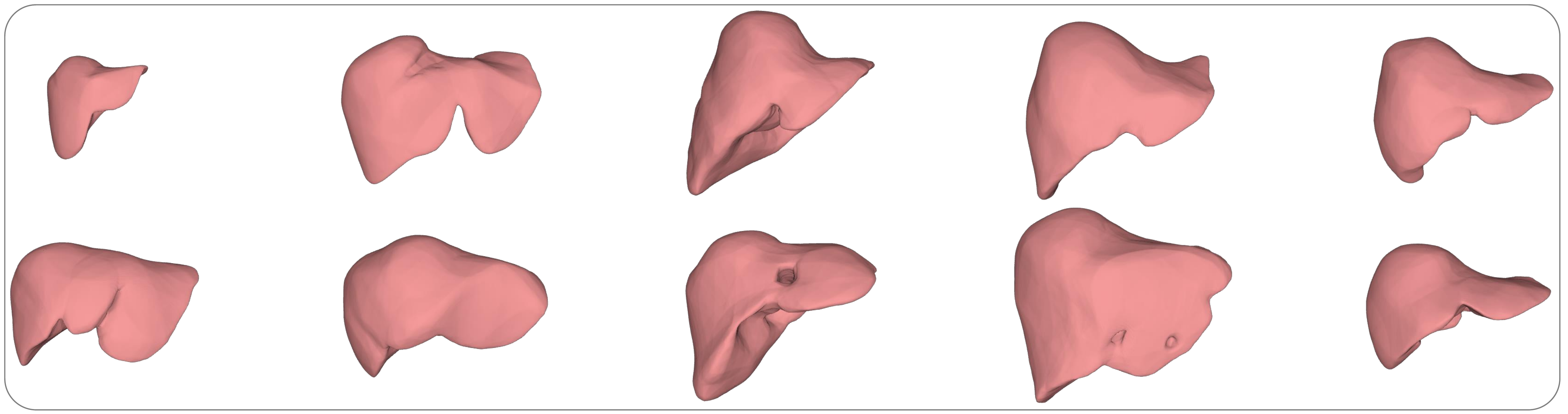}
        \caption{Visualization of synthesized 3D liver objects}
        \label{fig:overview_4}
    \end{subfigure}
    \caption{
   Overview of the proposed 3D liver HyperDiffusion framework.  
    Our framework consists of two training stages: (a) MLP training and (b) HyperDiffusion on MLP training. 
    (c) To synthesize a novel 3D liver object, we use the optimized HyperDiffusion network to sample the novel $\theta$ from the diffusion process.  
    The novel $\theta$ is reshaped back to an MLP and used to reconstruct the novel 3D liver object through the Marching Cubes (MC) algorithm.  
    (d) Visualization of the novel 3D liver objects.
    }
    \label{fig:overview}
\end{figure}

We adopt the framework of HyperDiffusion~\cite{erkocc2023hyperdiffusion} as a 3D liver generative model to generate additional synthetic 3D objects in the form of INRs.  
The framework consists of two stages: 
(1) training MLPs to represent each 3D liver object as an INR and
(2) training a HyperDiffusion model in the space of MLP weights.

Once the HyperDiffusion model is trained, new MLP weights corresponding to valid INRs can be synthesized through the reverse diffusion process from a randomly sampled noise signal.
The generated 3D liver objects are then reconstructed using the Marching Cubes (MC) algorithm\footnote{\texttt{marching\_cubes} from \href{https://scikit-image.org/docs/stable/api/skimage.measure.html}{scikit-image}}\cite{lorensen1998marching}.
Fig.~\ref{fig:overview} illustrates our workflow.

\begin{figure}[t]
    \centering
    \includegraphics[width=1\linewidth]{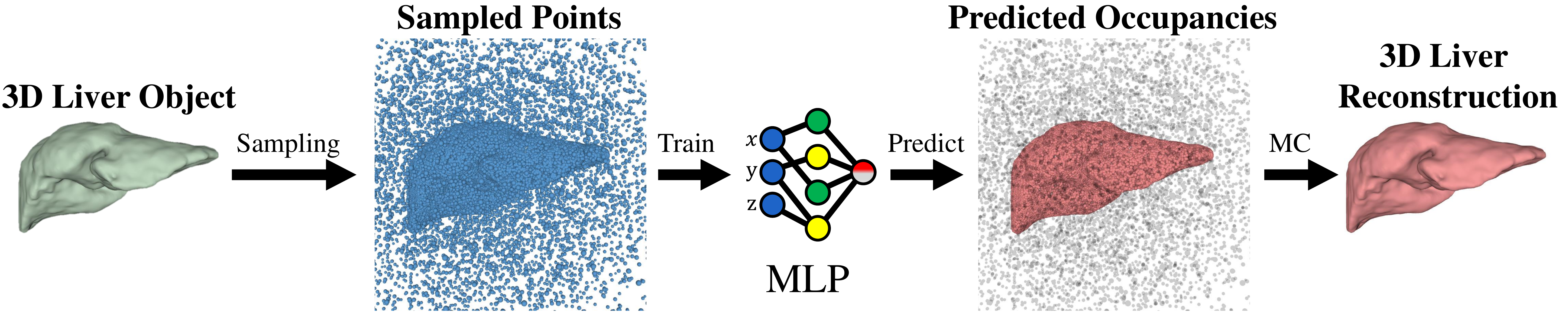}
    \caption{
    Our workflow used for MLP training.  
    From the 3D liver object, we sample a point cloud both inside and outside the object's surface.  
    Given the coordinates of each point as input, the MLP predicts the occupancy value: $1$ for inside (\textcolor{red}{red} point color) and $0$ for outside (\textcolor{gray}{gray} point color).  
    To reconstruct the 3D liver object from the volume of occupancy values, we use the MC algorithm.
    }
    \label{fig:inr}
\end{figure}

\subsection{Instance Liver MLP}

In the first stage, for each surface $\mathbf{S}_i$ of a 3D object in the training set $\{\mathbf{S}_i | i= 1,...,N\}$, we train a MLP to predict the occupancy values by minimizing the binary cross-entropy loss function:
\begin{equation}
L= \text{BCE}(f(\mathbf{x}, \theta_i), \mathbf{o}_i(x))~\text{for}~\mathbf{x} \in \mathbb{R}^3,
\end{equation}
where $f$ is a MLP parametrized by $\theta_i \in \mathbb{R}$, $\mathbf{x}$ represents a 3D coordinate and $\mathbf{o}_i(x) \in \{0,1\}$ is the ground-truth occupancy of $x$ ($1$ for inside and $0$ for outside) with respect to $\mathbf{S}_i$.

\textbf{MLP Architecture.}
The MLP architecture is a regular fully connected network, where the input is a 3D coordinate concatenated with a positional encoding~\cite{mildenhall2020nerf}. 
It consists of three hidden layers, each with 128 neurons and ReLU activation functions.
The output is a scalar occupancy value~\cite{Occupancy_Networks}.
In addition, we use the same MLP architecture for different 3D objects in the training dataset.
However, each MLP is optimized independently for a single 3D object, meaning there is no parameter sharing among the training set.

\textbf{MLP Training.}
To train the MLP, we normalize all 3D objects to fit within the volume range $[-0.5, 0.5]^3$ and randomly sample $20k$ points within this volume.
To enhance surface precision at a finer scale, we additionally sample another $20k$ points near the surface, resulting in a total of $40k$ sampled points.  
To distinguish whether a point is inside or outside the surface, we compute the generalized winding numbers\footnote{\texttt{fast\_winding\_number\_for\_meshes} from \href{https://libigl.github.io/libigl-python-bindings/igl_docs}{libigl}}\cite{barill2018fast} to obtain the ground-truth occupancy values.
We train each MLP for $1000$ epochs with a mini-batch size of $2048$ points per epoch, which takes approximately 1 minute per 3D object.
To reconstruct the 3D liver object, we apply the MC algorithm to extract the surface from the predicted occupancy values.
The workflow is illustrated in Fig.~\ref{fig:inr}.
All of the optimized MLPs will serve as the training set for the next stage.

\begin{figure}[t]
    \centering
    \includegraphics[width=1\linewidth]{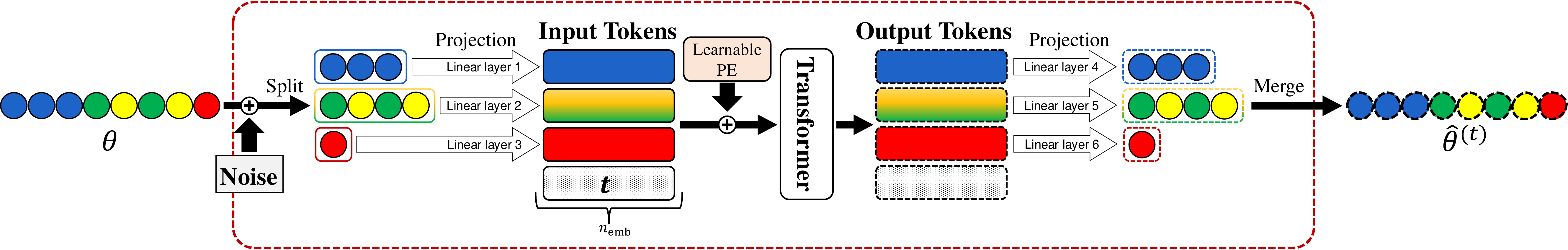}
    \caption{
    The operation in HyperDiffusion during training.  
    The flattened $\theta$ is added random Gaussian noise and split into a sequence of tokens through projection.
    An additional time step token is appended to the sequence to indicate the current time step $t$, and the sequence is then added to a learnable Positional Encoding (PE) before being input into the Transformer.  
    The output tokens are projected back to the size of the weights and biases to merge and represent the denoised weights $\hat{\theta}^{(t)}$.
    }
    \label{fig:hyperdiffusion}
\end{figure}

\subsection{HyperDiffusion on Liver MLPs}
In the second stage, we train a diffusion model on the optimized MLP weights and biases $\{\theta_i\}_{i=1}^{N}$ to model the space of weights through the diffusion process.
We use the Transformer-based diffusion model from G.pt~\cite{Peebles2022} and HyperDiffusion~\cite{erkocc2023hyperdiffusion} as the baseline for our HyperDiffusion on liver MLPs, as shown in Fig.~\ref{fig:hyperdiffusion}. 

\textbf{HyperDiffusion Training.} 
Given an MLP as input into HyperDiffusion, we flatten it into a 1D vector $\theta$ and add Gaussian noise at time step $t\in[1,T=1000]$ as part of the forward diffusion process.  
The noisy $\theta$ is then split into a sequence of tokens, where the sequence length corresponds to the number of weight and bias tensors in $\theta$:  
$[3456, 128, 16384, 128, 16384, 128, 128, 1]$
resulting in a sequence length of 8, according to our MLP design.  
Since each weight and bias tensor has a different size, a projection step is required to map them all to a uniform embedding size of $n_{\text{emb}} = 2880$. 
Each weight and bias tensor has a corresponding linear projection layer, and these layers are not shared among them.
A sinusoidal embedding of the time step $t$ is concatenated as an additional token, and all the input tokens are then combined with a learnable Positional Encoding (PE).  
The output tokens are then projected back to the original size and merged into a 1D vector, representing the denoised weights $\hat{\theta}^{(t)}$ at time step $t$.
The Mean Squared Error (MSE) loss is calculated between the denoised weights $\hat{\theta}^{(t)}$ and the original weights $\theta$.
The MSE loss guides the model to minimize the difference between the denoised weights and the true weights during the reverse diffusion process.

\textbf{Synthesizing New 3D Liver Objects.}
After training, we use Denoising Diffusion Implicit Models (DDIM)~\cite{songdenoising} to sample new MLPs. 
Starting with a pure noise $\theta$ (initialized as random Gaussian noise), the DDIM process gradually denoises it over several steps, progressively moving it towards a valid MLP that represents a new 3D liver object.

%% file: chapters/4_Experiment.tex
\section{Experiments}

\textbf{Setup.}
Recalling Fig.~\ref{fig:TotalSegmentator} from Sec.~\ref{sec:data_analysis}, we use 452 usable 3D objects (`Usable') as our liver dataset.
We first train MLPs on the entire dataset 
to obtain a collection of 452 liver MLPs. 
These MLPs are then split into non-overlapping partitions: training ($80\%$), validation ($5\%$), and testing ($15\%$) subsets for HyperDiffusion.  
We train HyperDiffusion for $6000$ epochs using the AdamW optimizer with a batch size of $32$ and a learning rate of $2e^{-4}$ on a single A6000 GPU, which takes approximately $9$ hours.
We also attempted to train a 1D UNet-based model\footnote{from \url{https://github.com/lucidrains/denoising-diffusion-pytorch}} to compare its performance with the transformer-based model by replacing the Transformer in HyperDiffusion with a 1D UNet.
However, it takes approximately $22$ hours to converge after $12000$ epochs.

\begin{table}[tb]
    \centering
    \caption{
    Quantitative results of liver MLPs on 452 3D liver objects (`Usable'). 
    We use the following evaluation metrics: VIoU, NC, F-Score (higher is better), and Chamfer-$L_1$ distance (lower is better). 
    The scores across all metrics indicate that our MLPs effectively represent the 3D liver objects, demonstrating the quality of the learned representations.
    }
    \label{tab:mlp}
    \begin{tabularx}{\linewidth}{*4{>{\centering\arraybackslash}X}}
        \hline
        VIoU~$\uparrow$ & Chamfer-$L_1$~$\downarrow$ & NC~$\uparrow$ & F-Score~$\uparrow$ \\ \hline
        0.9747 & 0.0028 & 0.9780 & 1.0000 \\ \hline
    \end{tabularx}
\end{table}

\begin{table}[t]
    \centering
    \caption{
    We scale the MMD by $10^2$, while COV and 1-NNA are measured in percentages ($\%$). 
    For these metrics, higher values of COV are preferred, while lower values of MMD, 1-NNA, and FPD are desirable. 
    Specifically, an optimal 1-NNA value is $50\%$.
    Overall, the Transformer-based model outperforms the 1D UNet-based model across all metrics.
    }
    \label{tab:hyperdiffusion}
    \begin{tabularx}{\linewidth}{*5{>{\centering\arraybackslash}X}}
        \hline
        Model Type & MMD~$\downarrow$ & COV (\%)~$\uparrow$ & 1-NNA (\%)~$\downarrow$ & FPD~$\downarrow$ \\ \hline
        Transformer & \textbf{0.24} & \textbf{55.88} & \textbf{53.68} & \textbf{2.27} \\
        1D UNet & 0.25 & 42.65 & 68.38 & 6.41 \\
        \hline
    \end{tabularx}
\end{table}

\noindent\textbf{Evaluation Metrics.}
We follow~\cite{knapitsch2017tanks,Occupancy_Networks,Peng2021SAP} to evaluate the quality of 3D reconstructions from liver MLPs using the following metrics: Volumetric Intersection over Union (VIoU), Chamfer-$L_1$ distance, Normal Consistency (NC), and F-Score with a default threshold of $1\%$.
The quantitative results in Table~\ref{tab:mlp} show that our liver MLPs achieve high scores, nearly $1$, in VIoU and NC. 
Additionally, we achieve the highest score in F-Score (with a maximum value of $1$) and a Chamfer-$L_1$ distance close to zero.
These results demonstrate that our liver MLPs effectively represent the 3D liver objects, providing a high-quality dataset for training HyperDiffusion. 

To evaluate the quality of the reconstructed 3D liver object synthesis, we use the metrics from HyperDiffusion~\cite{erkocc2023hyperdiffusion}: Minimum Matching Distance (MMD), Coverage (COV), and 1-Nearest-Neighbor Accuracy (1-NNA). 
Since there is no specific ground truth for the synthesis, these metrics help assess how well the synthesized 3D objects match the existing real 3D liver objects. 
We measure the 3D liver object synthesis using the test set to evaluate how well the model generalizes to unseen data. 
The results in Table~\ref{tab:hyperdiffusion} show that the Transformer-based model is better than the 1D UNet-based model, with FPD being lower by a factor of 2.8 and 1-NNA closer to $50\%$.
The synthesized 3D liver objects from the Transformer-based model will be used in our next assessment.

\begin{figure}[t]
    \centering
    \begin{subfigure}[b]{0.62\textwidth}
        \centering
        \includegraphics[width=\textwidth]{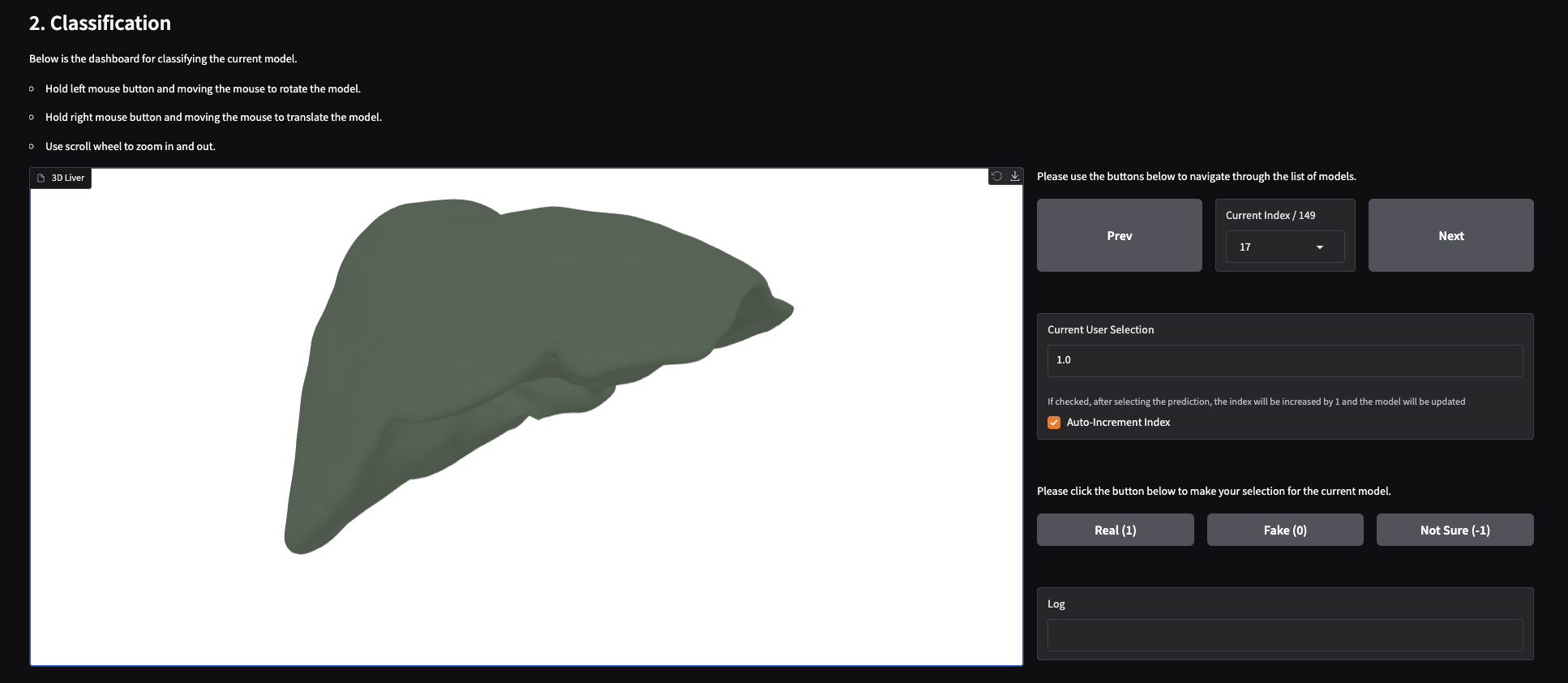}
        \caption{Survey GUI}
        \label{fig:survey_1}
    \end{subfigure}
    \hfill
    \begin{subfigure}[b]{0.36\textwidth}
        \centering
        \includegraphics[width=\textwidth]{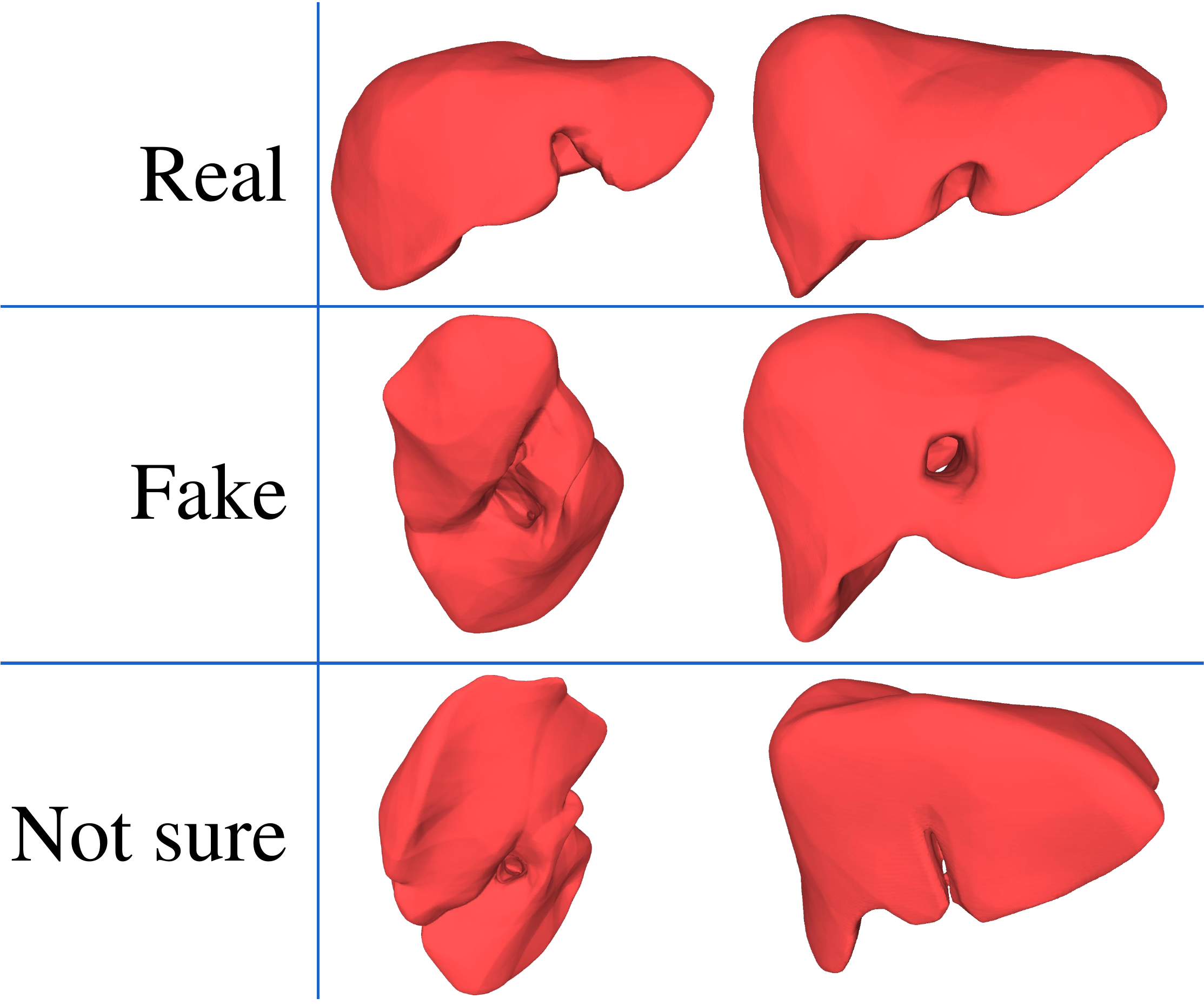}
        \caption{Visualization}
        \label{fig:survey_2}
    \end{subfigure}
    \caption{
    (a) The GUI allows the user to view one 3D liver object at a time and navigate among 150 objects to review and select from the options -- Real, Fake, and Not sure -- to classify the current 3D liver object.  
    (b) Visualization of two synthesized 3D liver objects for each type of classification.
    }
    \label{fig:survey}
\end{figure}

\noindent\textbf{Expert Evaluation of 3D Liver Objects.}
We conducted an additional survey to evaluate our synthesized 3D liver objects at an expert level. 
We asked our surgical team 
to classify a collection of 150 3D liver objects, consisting of $50\%$ real liver objects (68 from the test set and 7 from the validation set) and $50\%$ fake liver objects (75 synthesized liver objects).
As shown in Fig.~\ref{fig:survey}, 
there are three options -- Real, Fake, and Not sure -- to classify all 150 3D liver objects.

After completing the survey, the selections are: Real (139), Fake (4), and `Not sure' (7). 
According to the members of the surgical team, they classified a liver as real if the overall structure is well represented, even with small holes on the posterior face, and selected `Not sure' only when the holes are too prominent and deform the shape.  
Based on the survey results, our synthesized 3D liver objects appear realistic, as all were classified as real.
However, some cases contain arbitrary holes throughout the structure, as labeled `Fake' in Fig.~\ref{fig:survey_2}, but the overall structure is well synthesized by our 3D Liver HyperDiffusion model.


%% file: chapters/5_Conclusions.tex
\section{Conclusions}
In this paper, we presented our findings on the lack of high-quality data in 3D medical shape datasets and proposed using the HyperDiffusion model to generate synthetic 3D liver objects to address this issue.  
Our results demonstrated that the quality of the synthesized 3D liver objects aligns well with the real liver structure.  
In future work, we plan to incorporate conditional methods to control the synthesized shapes, such as text-to-3D or image-to-3D generation.